\documentclass[10pt,twocolumn,letterpaper]{article}

\usepackage[T1]{fontenc}
\usepackage[utf8]{inputenc}

\usepackage{cvpr}              

\usepackage{algorithm}
\usepackage{algpseudocode}
\definecolor{cvprblue}{rgb}{0.21,0.49,0.74}
\usepackage[pagebackref,breaklinks,colorlinks,allcolors=cvprblue]{hyperref}

\def\paperID{}  
\def\confName{CVPR}
\def\confYear{2026}

\title{Generative Anonymization in Event Streams}

\author{Adam T. Müller \quad Mihai Kocsis \quad Nicolaj C. Stache\\
Heilbronn University of Applied Sciences, Germany\\
{\tt\small \{adam-theo.mueller, mihai.kocsis, nicolaj.stache\}@hs-heilbronn.de}
}

\begin{document}
\maketitle
\begin{abstract}
Neuromorphic vision sensors offer low latency and high dynamic range, but their deployment in public spaces raises severe data protection concerns. Recent Event-to-Video (E2V) models can reconstruct high-fidelity intensity images from sparse event streams, inadvertently exposing human identities. Current obfuscation methods, such as masking or scrambling, corrupt the spatio-temporal structure, severely degrading data utility for downstream perception tasks.

\noindent In this paper, to the best of our knowledge, we present the first generative anonymization framework for event streams to resolve this utility-privacy trade-off. By bridging the modality gap between asynchronous events and standard spatial generative models, our pipeline projects events into an intermediate intensity representation, leverages pretrained models to synthesize realistic, non-existent identities, and re-encodes the features back into the neuromorphic domain. Experiments demonstrate that our method reliably prevents identity recovery from E2V reconstructions while preserving the structural data integrity required for downstream vision tasks. Finally, to facilitate rigorous evaluation, we introduce a novel, synchronized real-world event and RGB dataset captured via precise robotic trajectories, providing a robust benchmark for future research in privacy-preserving neuromorphic vision.

\end{abstract}

\section{Introduction}
\label{sec:intro}

Neuromorphic vision sensors, or event cameras, represent a paradigm shift in visual perception. Unlike conventional cameras that capture dense, synchronous frames at fixed intervals, event cameras respond asynchronously to per-pixel changes in illumination~\cite{Hu2021-v2e-cvpr-workshop-eventvision2021}. This fundamental difference enables sensors with microsecond latency, high dynamic range, and minimal power consumption, making them highly attractive for highly dynamic real-world applications such as autonomous driving, robotics, and smart surveillance~\cite{chen_event-based_2020,kaminski_observational_2019,zhou_deep_2024}. However, as the deployment of these sensors accelerates in public and human-centric spaces, the treatment of the data they capture falls under stringent data protection regulations~\cite{eu_ai_act}.

Because event streams inherently lack absolute intensity information and only encode scene dynamics as a sparse point cloud of brightness changes, there has been a passive assumption that they do not capture sensitive biometric information~\cite{ahmad_event-driven_2022,becattini_understanding_2022}. Recent advancements in deep learning have demonstrated that this assumption is flawed, as state-of-the-art event-to-video (E2V)~\cite{ercan_evreal_2023,qu_e2hqv_2024} reconstruction models are able to recover high-fidelity intensity images from raw event streams. Consequently, an unprotected event stream capturing biometric identifiers can easily be inverted to reveal the subject's identity, introducing a severe privacy vulnerability.

To mitigate this risk, early works in event privacy have proposed pertubation based approaches~\cite{ahmad_event_2024,bendig_anonynoise_2025}. While effective against reconstruction attacks, such anonymization methods inherently corrupt the underlying spatio-temporal structure of the event stream, severely limiting the utility of the anonymized data for downstream tasks. In the conventional frame-based domain, this utility-privacy trade-off is being addressed through \textit{generative anonymization}, utilizing advanced machine learning based models to seamlessly replace a person's identity with a synthetic, non-existent one while preserving semantic context, gaze, and pose~\cite{klemp_ldfa_2023,zwick_context-aware_2024}.

In this paper, we present the first step toward bridging this utility-privacy gap by introducing generative identity anonymization to the event domain. We propose a framework that maps the principles of generative RGB anonymization into the event space. Our approach utilizes the maturity of RGB-based pretrained face-swapping models. We detect and generatively anonymize facial identities in projected frame-space, and subsequently utilize robust video-to-event (V2E) processing to project these synthesized identities into an anonymized event stream. When subjected to E2V reconstruction attacks, our processed streams yield realistic human features belonging to a newly generated identity, effectively protecting the original subject without destroying the structural integrity of the data.

Furthermore, research in event-based human perception is restricted by a lack of high-quality, real-world datasets. Existing collections, such as the FES dataset by ISSAI~\cite{bissarinova_faces_2024}, frequently rely on synthetic video-to-event (V2E) generation or feature uncontrolled subject movement. To facilitate rigorous evaluation and future research, we introduce a novel dataset featuring synchronized real-world event and RGB streams. To ensure precise, reproducible egomotion and a static subject setup, the sensor suite was mounted on a collaborative robot (cobot) executing programmed trajectories.

In summary, our main contributions are as follows: \textbf{(1)} We propose a information retaining generative anonymization pipeline for event streams, utilizing frame-space diffusion models and V2E projection to synthesize realistic, alternative identities. \textbf{(2)} We demonstrate that the proposed anonymization reliably prevents identity recovery from high-fidelity E2V reconstructions while preserving the spatio-temporal utility necessary for downstream vision tasks. \textbf{(3)} We introduce a synchronized real-world event-RGB dataset\href{https://github.com/muelleradam/KinematicEvent-HumanUpperBody-2026}{}\footnote{https://github.com/muelleradam/KinematicEvent-HumanUpperBody-2026}, captured via precise cobot trajectories, to spur further research in privacy-preserving neuromorphic vision.

\section{Related Work}
\label{sec:relWork}

\paragraph{Event-to-Video and Image Reconstruction.}

Neuromorphic vision sensors encode visual information asynchronously as sparse streams of brightness changes~\cite{Hu2021-v2e-cvpr-workshop-eventvision2021}. To bridge the modality gap between this non-standard data and conventional computer vision pipelines, a rich body of work has focused on E2V~\cite{ercan_evreal_2023} reconstruction. Seminal learning-based approaches, such as E2VID~\cite{rebecq_events--video_2019}, successfully demonstrated the recovery of high frame-rate absolute intensity videos from event streams using recurrent neural networks. Subsequent advancements introduced lightweight, high-speed alternatives like FireNet~\cite{scheerlinck_fast_2020}, as well as high-fidelity models such as ET-Net~\cite{weng_event-based_2021}. While the field continues to rapidly advance with highly complex recent architectures~\cite{qu_e2hqv_2024,ge_event-based_2025,zou_eventhdr_2025}, deploying and standardizing these models across different sensor setups remains challenging. Consequently, we utilize the highly robust EVREAL evaluation framework~\cite{ercan_evreal_2023} for our pipeline.

Crucially, the continuous refinement of these E2V methodologies has inadvertently demonstrated the reliable extraction of highly identifiable facial details from sparse event spikes. These reconstruction networks have exposed a severe privacy vulnerability, directly motivating the necessity of data anonymization frameworks.

\paragraph{Privacy and Anonymization in Event Streams.}

As the capabilities of E2V reconstruction models matured, the privacy risks associated with neuromorphic sensors spurred pioneering works in the domain of event stream anonymization. To conceal identities, initial approaches employ techniques such as spatial scrambling, adversarial noise injection, or the direct encryption of event spikes~\cite{du_event_2021}. Most notable are the EventAnon frameworks introduced by Ahmad \etal~\cite{ahmad_event-driven_2022,ahmad_person_2023,ahmad_event_2024}, seeking to mitigate privacy vulnerabilities by proposing end-to-end architectures optimized jointly for identity obfuscation and specific macroscopic downstream tasks, such as person re-identification or pose estimation. Building upon this, Bendig et al. recently introduced AnonyNoise~\cite{bendig_anonynoise_2025}, which applies learnable, data-dependent noise to raw event streams, preventing neural network-based re-identification while retaining coarse information for downstream tasks.

While highly effective at thwarting facial reconstruction via re-identification networks, these methodologies fundamentally degrade information by design. By intentionally obfuscating or displacing precise event coordinates, such methods inherently corrupt the localized spatio-temporal structure of the raw data. Such data degradation introduces a severe utility-privacy trade-off, bottlenecking the performance of fine-grained perception tasks that rely on high-fidelity structural integrity, such as facial expression recognition~\cite{berlincioni_neuromorphic_2023} or dense tracking~\cite{gehrig_dense_2024}.

Taking a different approach, Adra and Dugelay proposed E2PRIV~\cite{adra_e2priv_2025}, shifting the obfuscation step by integrating distortion based anonymization directly into the E2V reconstruction process. However, as E2PRIV avoids altering the raw event stream, it provides no privacy protection against malicious actors or perception networks operating directly in the event-space.

To address privacy natively in the event-space, without incurring to the limitations of destructive obfuscation, our work proposes a paradigm shift toward generative anonymization. By seamlessly replacing sensitive features with synthesized, non-existent identities while fully preserving the underlying utility of the event stream.

\paragraph{Generative Anonymization in RGB Images.}

In contrast to the destructive obfuscation techniques currently utilized in event-based vision, the utility-privacy trade-off is resolved through generative anonymization in the frame-based domain. Early models in this area used Generative Adversarial Networks (GANs), such as DeepPrivacy~\cite{hukkelas_deepprivacy_2019} and CIAGAN~\cite{maximov_ciagan_2020}, to synthesize photorealistic, non-existent faces that replace sensitive identities while retaining critical semantic attributes like head pose and gaze.

Recently, the field has experienced a paradigm shift driven by Latent Diffusion Models (LDMs), which offer enhanced image fidelity and context-awareness. Notably, Klemp \etal~\cite{klemp_ldfa_2023} introduced LDFA, a pipeline based on stable diffusion to perform seamless, context-aware facial anonymization for autonomous driving datasets. Building upon this foundation, Zwick \etal~\cite{zwick_context-aware_2024} extended this generative approach to the entire human body, effectively removing secondary biometric identifiers such as clothing and posture while maintaining the structural semantics of the scene. However, these powerful generative priors are strictly designed for dense, synchronous spatial tensors and cannot natively ingest the asynchronous, sparse nature of raw event streams~\cite{yang_gface_2024}.

Our work aims to bridge this modality gap, bringing the idea of generative anonymization to event data. We propose a pipeline that projects event data into the continuous intensity domain, such that methods from the frame-based domain can be applied for anonymization, and subsequently re-encode the synthesized identities back into the neuromorphic space.

\section{Method}
\label{sec:Method}

\begin{figure*}[t]
  \centering
   \includegraphics[width=\textwidth]{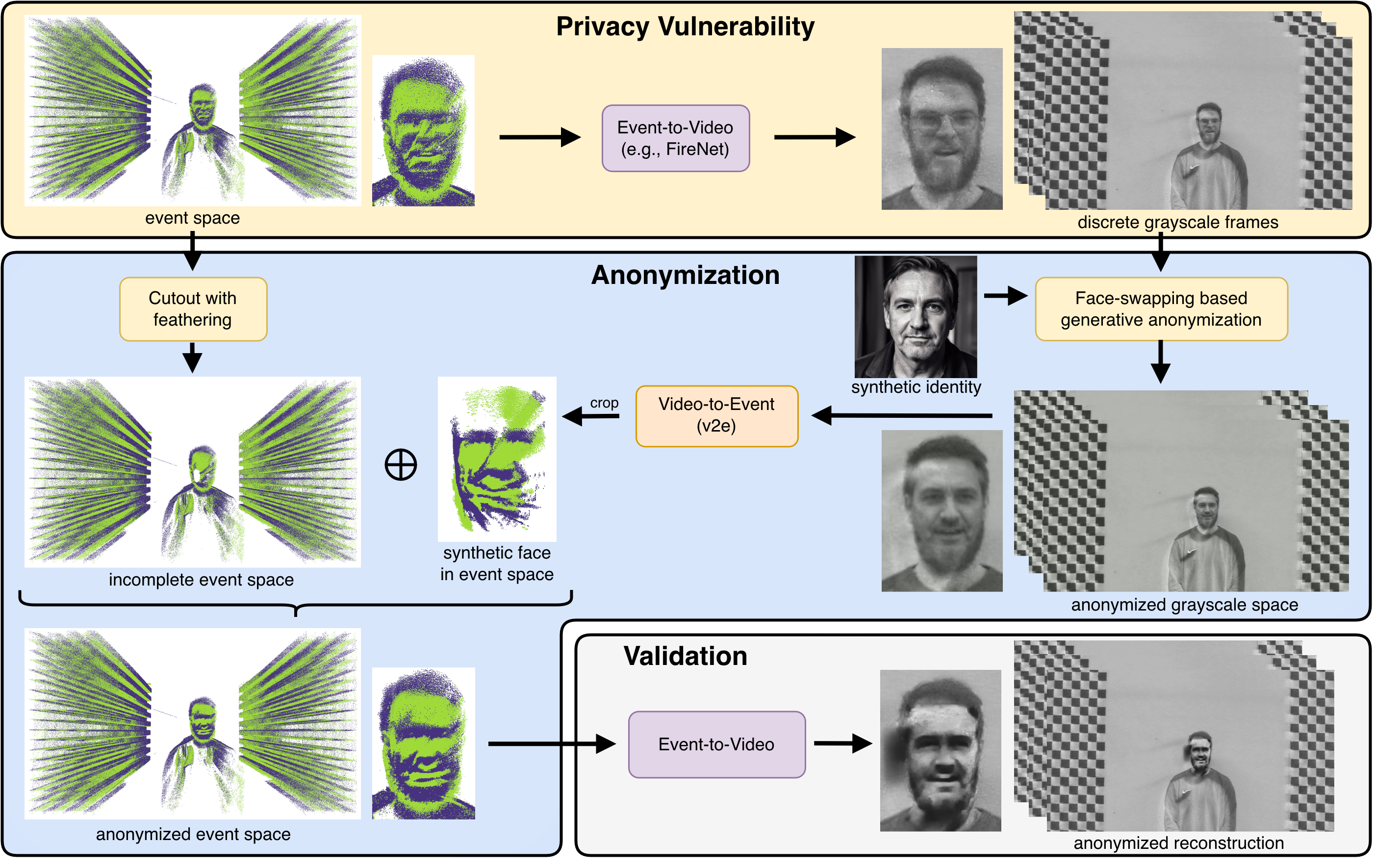}
   \caption{\textbf{Architectural overview of the generative anonymization pipeline.} The framework translates raw asynchronous event data into continuous grayscale frames to detect and swap faces using established generative models. The anonymized identity is subsequently projected back into the event space via a V2E conversion, preserving the underlying spatiotemporal structure.}
   \label{fig:EventAnon}
\end{figure*}

To achieve high-fidelity generative anonymization in the neuromorphic domain, we must navigate the incompatibility between asynchronous event streams and standard spatial generative models. In our proposed framework, we address this by bridging the modality gap via intermediate intensity representations.

\subsection{Generative Anonymization}
The full workflow is shown in~\cref{fig:EventAnon}. As an initial step, we translate the raw event data into the frame-based grayscale space. Let the asynchronous event stream be defined as a set of events $\mathcal{E}$:

\begin{equation}
    \mathcal{E} = \{e_i\}_{i=1}^{N},
\end{equation}

\noindent where each event $e_i = (x_i, y_i, t_i, p_i)$ consists of its spatial pixel coordinates $(x_i, y_i)$, a microsecond-resolution timestamp $t_i$, and the polarity $p_i \in \{-1, +1\}$. To project this data into the continuous intensity domain, we employ methods within the EVREAL evaluation framework. This process converts the sparse event stream into a set of $K$ synchronized intensity frames.

\paragraph{Data Anonymization.} We first detect the face in a given frame $k$, where the bounding box at the $k$-th frame, captured at time $T_k$, be defined by its top-left and bottom-right spatial coordinates:
\begin{equation}
    B_k = (x_{1,k}, y_{1,k}, x_{2,k}, y_{2,k}).
    \label{eq:BboxDef}
\end{equation}

\noindent We then utilize the open INSwapper~\cite{deng_arcface_2022,chen_simswap_2020} face swapping model to replace the subject with a new identity. For this we use Stable Diffusion 2 (SD2)~\cite{rombach_high-resolution_2022} to generate a new synthetic identity as an input for INSwapper. We then translate the anonymized grayscale data back into the event space using v2e~\cite{Hu2021-v2e-cvpr-workshop-eventvision2021}.

\paragraph{Temporal Interpolation of Spatial Boundaries.} We can use the bounding box information in~\cref{eq:BboxDef} to cut out the non-anonymized subject (face) from the baseline event stream and crop the output event stream from v2e to only contain the synthesized face.

Because event cameras have microsecond temporal resolution, the discrete bounding boxes $B_k$ must be interpolated to evaluate the spatial limits at the exact time $t_i$ of any arbitrary event. We compute a continuous bounding box function $B(t)$ using 1D piecewise linear interpolation.

For any event occurring at time $t_i$ such that $T_k \leq t_i < T_{k+1}$, the interpolated top-left $x$-coordinate, $x_1(t_i)$, is defined as:
\begin{equation}
    x_1(t_i) = x_{1,k} + \frac{x_{1,k+1} - x_{1,k}}{T_{k+1} - T_k} (t_i - T_k).
\end{equation}

\noindent This same linear interpolation is applied independently to compute $y_1(t_i)$, $x_2(t_i)$, and $y_2(t_i)$. This establishes a dynamic, continuously moving bounding box in the 3D spatiotemporal volume:
\begin{equation}
    B(t) = (x_1(t), y_1(t), x_2(t), y_2(t)).
\end{equation}

\noindent Finally, the event stream is filtered to extract only the events that fall within this dynamic Region of Interest (ROI). An event $e_i$ is considered inside the bounding box if its spatial coordinates satisfy:
\begin{equation}
    x_1(t_i) \leq x_i \leq x_2(t_i) \quad \text{and} \quad y_1(t_i) \leq y_i \leq y_2(t_i).
\end{equation}

\noindent The resulting cropped event subset $\mathcal{E}_{\text{ROI}}$ is therefore defined as:
\begin{equation}
\begin{split}
    \mathcal{E}_{\text{ROI}} = \{ e_i \in \mathcal{E} \mid & x_1(t_i) \leq x_i \leq x_2(t_i) \\
    & \land y_1(t_i) \leq y_i \leq y_2(t_i) \}.
\end{split}
\end{equation}

\noindent This formulation satisfies the condition for the crop to the facial region of the anonymized event stream. For the cutout of the background data in the baseline stream, the logic has to be changed to $e_i \in \mathcal{E} \backslash \mathcal{E}_{ROI}$.

\paragraph{Stochastic Spatial Feathering.} To mitigate harsh artificial boundaries when extracting the background event stream (i.e., the cutout $\mathcal{E} \setminus \mathcal{E}_{\text{ROI}}$), we introduce a spatial feathering mechanism inspired by Gaussian blending~\cite{burt_multiresolution_1983} and probabilistic event sampling strategies~\cite{girbau-xalabarder_probabilistic_2025}. Rather than employing a strict binary spatial threshold, events located strictly inside the bounding box are retained with a probability that decays according to a half-Gaussian distribution. Let $d(e_i, \partial B(t_i))$ denote the shortest Euclidean distance from the spatial coordinates of an event $e_i$ to the perimeter of the bounding box $\partial B(t_i)$ at time $t_i$. The feathered background stream, $\mathcal{E}_{\text{bg}}$, is generated by sampling events from $\mathcal{E}$ according to the retention probability:
\begin{equation}
    P(e_i \in \mathcal{E}_{\text{bg}}) = 
    \begin{cases} 
        1, & \text{if } e_i \notin \mathcal{E}_{\text{ROI}} \\ 
        \exp\left(-\frac{d(e_i, \partial B(t_i))^2}{2\sigma^2}\right), & \text{if } e_i \in \mathcal{E}_{\text{ROI}},
    \end{cases}
\end{equation}
\noindent where the hyperparameter $\sigma$ controls the standard deviation, and thus the spatial width of the blending overlap region. This formulation ensures a smooth, probabilistic gradient of event density bridging the preserved background and the masked region.

\paragraph{Spatiotemporal Alignment and Stream Compositing.} To composite the extracted filler stream $\mathcal{E}_{\text{anon}}$ into the target background stream $\mathcal{E}_{\text{bg}}$, we apply a continuous spatiotemporal transformation to align both the timestamps and the spatial coordinates. 

To spatially warp the anonymized events into the target region of interest, we compute a dynamic, center-relative affine mapping based on the interpolated bounding box trajectories. Let $c_{\text{anon}}(t) = (c_{\text{anon},x}(t), c_{\text{anon},y}(t))$ and $c_{\text{tgt}}(t)$ denote the continuous spatial centers of the bounding boxes of both the anonymized and the background event stream at time $t$, with corresponding widths $w(t)$ and heights $h(t)$. For each anonymized event $e_i \in \mathcal{E}_{\text{anon}}$, the transformed spatial coordinates $(x'_i, y'_i)$ are mapped proportionally to the target bounding box via:
\begin{equation}
    x'_i = c_{\text{tgt},x}(t_i) + \left( \frac{x_i - c_{\text{anon},x}(t_i)}{w_{\text{anon}}(t_i)} \right) w_{\text{tgt}}(t_i),
\end{equation}
\begin{equation}
    y'_i = c_{\text{tgt},y}(t_i) + \left( \frac{y_i - c_{\text{anon},y}(t_i)}{h_{\text{anon}}(t_i)} \right) h_{\text{tgt}}(t_i).
\end{equation}

\noindent Finally, the mapped filler events are aggregated with the background stream, and the composite stream is chronologically sorted to produce the final synthetic event stream $\mathcal{E}_{\text{final}} = \mathcal{E}_{\text{bg}} \cup \mathcal{E}_{\text{anon}}'$.

\subsection{Evaluation}

To comprehensively evaluate the efficacy of the generative anonymization framework, we establish a robust set of quantitative metrics applied to the intermediate intensity representations. These metrics are specifically designed to assess both the strength of the identity obfuscation and the structural preservation of essential spatial and temporal features required for downstream utility.

\begin{itemize}
    \item \textbf{Identity Similarity.} To quantify the strength of the anonymization, we measure the distance between the identity embeddings of the source and the generatively anonymized data. Following the methodology of Egin \etal~\cite{egin_now_2025}, we compute the cosine similarity between feature vectors extracted using the ArcFace~\cite{deng_arcface_2022} network:

    \begin{equation}
     \text{Similarity}_{\text{ID}} = \frac{\boldsymbol{v}_{\text{src}} \cdot \boldsymbol{v}_{\text{gen}}}{||\boldsymbol{v}_{\text{src}}|| \cdot ||\boldsymbol{v}_{\text{gen}}||},
    \end{equation}

    where $\boldsymbol{v}_{\text{src}}$ and $\boldsymbol{v}_{\text{gen}}$ represent the 512-dimensional identity embeddings ($\boldsymbol{v} \in \mathbb{R}^{512}$) of the baseline and synthetic faces, respectively. An effective anonymization yields a low identity similarity score.

    \item \textbf{Temporal Stability.} To verify that the synthesized identity remains consistent across the duration of the event stream, we evaluate the frame-to-frame identity similarity within the anonymized data. The temporal stability, denoted as $\mathcal{S}_{\text{temp}}$, is defined as the average cosine similarity of embeddings between consecutive time steps:

    \begin{equation}
    \mathcal{S}_{\text{temp}} = \frac{1}{T-1} \sum_{t=1}^{T-1} \text{Similarity}_{\text{ID}}(\boldsymbol{v}_{t}, \boldsymbol{v}_{t+1}),
    \end{equation}

    where $\boldsymbol{v}_{t}$ and $\boldsymbol{v}_{t+1}$ are the identity embeddings extracted at frames $f_t$, $f_{t+1}$. An ideal temporal stability approaches 1.0, indicating a temporally coherent synthetic identity devoid of flickering or identity shifting.

    \item \textbf{Pose Error.} To ensure that critical geometric and behavioral semantics are retained post-anonymization, we measure the alignment between the head poses in the original and synthetic streams. Utilizing the HopeNet~\cite{ruiz_fine-grained_2018} architecture for head pose estimation, consistent with the approach by Ye \etal~\cite{ye_dreamid_2025}, we calculate the pose error $E_{\text{pose}}$ as the Mean Absolute Error (MAE) across the Euler angles:
    
    \begin{equation}
    E_{\text{pose}} = \frac{1}{3} (|y_{\text{orig}} - y_{\text{gen}}| + |p_{\text{orig}} - p_{\text{gen}}| + |r_{\text{orig}} - r_{\text{gen}}|)
    \end{equation}

    This metric quantifies the deviation in yaw ($y$), pitch ($p$), and roll ($r$), validating the preservation of the subject's spatial orientation.

    \item \textbf{Mimicry Error.} Similarly, to evaluate the preservation of fine-grained facial expressions, we compute the Landmark Distance (LMD) between the source and anonymized streams. Following Bulat \etal~\cite{bulat_how_2017}, we extract 106 2D facial keypoints, denoted as $\boldsymbol{L} \in \mathbb{R}^{106\times2}$, using the InsightFace detector~\cite{liu_grand_2019}. The mimicry error is thus formulated as the Euclidean distance between corresponding keypoints:

    \begin{equation}
    E_{\text{mimicry}} = \frac{1}{N} \sum_{i=1}^{N} \frac{|| \boldsymbol{L}_{\text{orig}}^{(i)} - \boldsymbol{L}_{\text{gen}}^{(i)} ||_2}{\text{IOD}},
    \end{equation}

    which is normalized by the Inter-Ocular Distance (IOD) to account for spatial scale variations across different subjects.

\end{itemize}

\vspace{0.13cm}
\noindent To evaluate the downstream utility of the anonymized event streams for practical computer vision tasks, we measure face detection performance across two distinct processing paradigms:

\begin{itemize}
    \item \textbf{Intensity-Domain Face Detection.} To assess performance in the frame-based intensity space, we utilize the YOLOv8 object detection architecture~\cite{redmon_you_2016,terven_comprehensive_2023}. We evaluate utility preservation through three primary metrics: (1) the mean detection confidence across all valid frames, (2) the spatial bounding box (BBox) shift, quantified by the Intersection over Union (IoU) between the baseline and anonymized detections, and (3) the relative error in overall detection rates, where an error of 0\% signifies perfectly consistent detection recall regardless of anonymization.

    \item \textbf{Event-Domain Face Detection.} To evaluate perception performance directly on the neuromorphic event-based data, we employ the pre-trained detection models introduced by Bissarinova \etal~\cite{bissarinova_faces_2024}. While this architecture relies on dense, accumulated event-frame representations rather than purely asynchronous raw spikes, it serves as a robust and representative benchmark for standard event-based vision pipelines. We measure the structural consistency of the anonymized stream by computing the BBox IoU on these event representations, ensuring that the critical spatiotemporal features required for event-based detection remain intact.

\end{itemize}

\vspace{0.13cm}
\noindent To rigorously assess the structural fidelity and spatiotemporal preservation of the anonymized data directly within the neuromorphic domain, we propose the following metrics for an evaluation of the raw event streams:

\begin{itemize}
    \item \textbf{Spatiotemporal Chamfer Distance (STCD).} To evaluate the strict structural preservation of the event streams, we employ a STCD~\cite{barrow_parametric_1977,fan_point_2017}:
    
\begin{equation}
\begin{split}
 d_{CD}(\mathcal{E}_1, \mathcal{E}_2) &= \frac{1}{|\mathcal{E}_1|} \sum_{e \in \mathcal{E}_1} \min_{e' \in \mathcal{E}_2} \|e - e'\| \\
 &\quad + \frac{1}{|\mathcal{E}_2|} \sum_{e' \in \mathcal{E}_2} \min_{e \in \mathcal{E}_1} \|e' - e\|
 \label{eq:chamfer_distance},
\end{split}
\end{equation}

    using $L_2$ Euclidean distance via KD-Tree search. Because event cameras generate asynchronous data, we treat the event streams as continuous 3D point clouds in space and time~\cite{sekikawa_eventnet_2019}. A critical challenge in computing spatial distances between events is the inherent scale mismatch between pixel coordinates and microsecond timestamps. To address this, we extract overlapping time windows of data and independently normalize the spatial (x, y) and temporal $t$ dimensions to a unit hypercube [0, 1]. This metric penalizes geometric distortions and the introduction of structural noise, yielding lower scores for similar spatiotemporal geometries.

    \item \textbf{Event Mover's Distance (EMD).} While Chamfer distance measures local nearest-neighbor similarity, it is insensitive to the overall density and global distribution of the events. As standard Earth Mover's Distance~\cite{rubner_earth_2000} solves an optimal transport problem that scales poorly to dense event streams, we approximate the true optimal transport cost between two spatiotemporal event sets $\mathcal{E}_1$ and $\mathcal{E}_2$ using the Sliced Wasserstein Distance (SWD)~\cite{bonneel_sliced_2015}. We project the event sets onto $L$ random unit vectors $\theta \in \mathbb{S}^{2}$ and compute the average of the 1D Wasserstein distances:

    \begin{equation}
     d_{EMD}(\mathcal{E}_1, \mathcal{E}_2) \approx d_{SW}(\mathcal{E}_1, \mathcal{E}_2) = \frac{1}{L} \sum_{l=1}^{L} W_1 \left( \mathcal{E}_1^{\theta_l}, \mathcal{E}_2^{\theta_l} \right),
     \label{eq:sliced_wasserstein}
    \end{equation}

    where $\mathcal{E}^{\theta_l}$ denotes the scalar projection of the event set onto the 1D line defined by $\theta_l$. For each 1D projection, the Wasserstein-1 distance $W_1$ is efficiently computed as the $L_1$ area between their empirical cumulative distribution functions $F_1$ and $F_2$:

    \begin{equation}
     W_1 \left( \mathcal{E}_1^{\theta_l}, \mathcal{E}_2^{\theta_l} \right) = \int_{-\infty}^{\infty} |F_1(x; \theta_l) - F_2(x; \theta_l)| \, dx.
     \label{eq:1d_wasserstein}
    \end{equation}

    This provides a computationally tractable, symmetric measure of the global distributional shift between two event streams.

\end{itemize}

\section{Experiments}
\label{sec:experiments}

To validate the proposed idea of generative anonymization of event streams and assess the practical utility of such processing, we present a study on quantitative anonymization measures as well as downstream task completion. Our evaluation focuses on three main objectives:

\begin{enumerate}[label=(\roman*)]
    \item \textbf{Effectiveness in Anonymization:} Establish the efficacy of the proposed anonymization.
    \item \textbf{Preservation of Features:} Demonstrate the usability of anonymized event data in downstream detection tasks.
    \item \textbf{Validity of Proposed Metrics:} Assess STCD and EMD as measures to evaluate anonymization of data streams in the event domain.
\end{enumerate}

\subsection{Experimental Setup} \label{subsec:ExpSetup}

For the initial E2V reconstruction, we employ the FireNet architecture, utilizing the implementation provided by the EVREAL~\cite{ercan_evreal_2023} framework. To perform generative anonymization in the intermediate frame space, we use the pre-trained INSwapper~\cite{chen_simswap_2020} 128 model, where the synthetic target identities are generated via SD2.

To mitigate the resolution constraints of the face-swapping prior, the anonymized outputs are upsampled by a factor of four using a Fast Super-Resolution Convolutional Neural Network (FSRCNN)~\cite{dong_accelerating_2016}. The spatial fidelity of these frames is subsequently refined through Contrast Limited Adaptive Histogram Equalization (CLAHE) and unsharp masking. Finally, the enhanced frame sequences are projected back into the neuromorphic domain using v2e~\cite{Hu2021-v2e-cvpr-workshop-eventvision2021}. For this conversion, the standard .h5 output format is utilized to support a maximum spatial resolution of 1024x768 pixels.

Crucially, empirical observations indicate that generating a sufficient event density during the V2E reverse projection step is vital. We note, that lack of localized event density translates into severe smearing artifacts and structural degradation when the anonymized stream is subjected to downstream E2V reconstruction.

\subsection{Event Face Dataset}

Existing event-based datasets, such as the corpus introduced by Berlincioni \etal~\cite{berlincioni_neuromorphic_2023}, are highly valuable but typically rely on subject movement to generate events. To eliminate human motion variance and better mimic applications where the camera is in motion (e.g., autonomous driving), we collected a novel, synchronized RGB-Event dataset utilizing a physical event sensor.

To decouple camera motion from human behavior, the sensor suite was mounted on a cobot executing programmed trajectories, guaranteeing precise egomotion. During capture, subjects were instructed to read short text passages. This specific task naturally induced facial micro-expressions and lip movements while preventing gross body motion, providing an ideal baseline for evaluating spatiotemporal structural integrity. This rigorously controlled, real-world dataset (availability see \cref{sec:intro}) provides a robust benchmark for evaluating generative event anonymization.

\subsection{Results}

\begin{figure}[t]
  \centering
   \includegraphics[width=\linewidth]{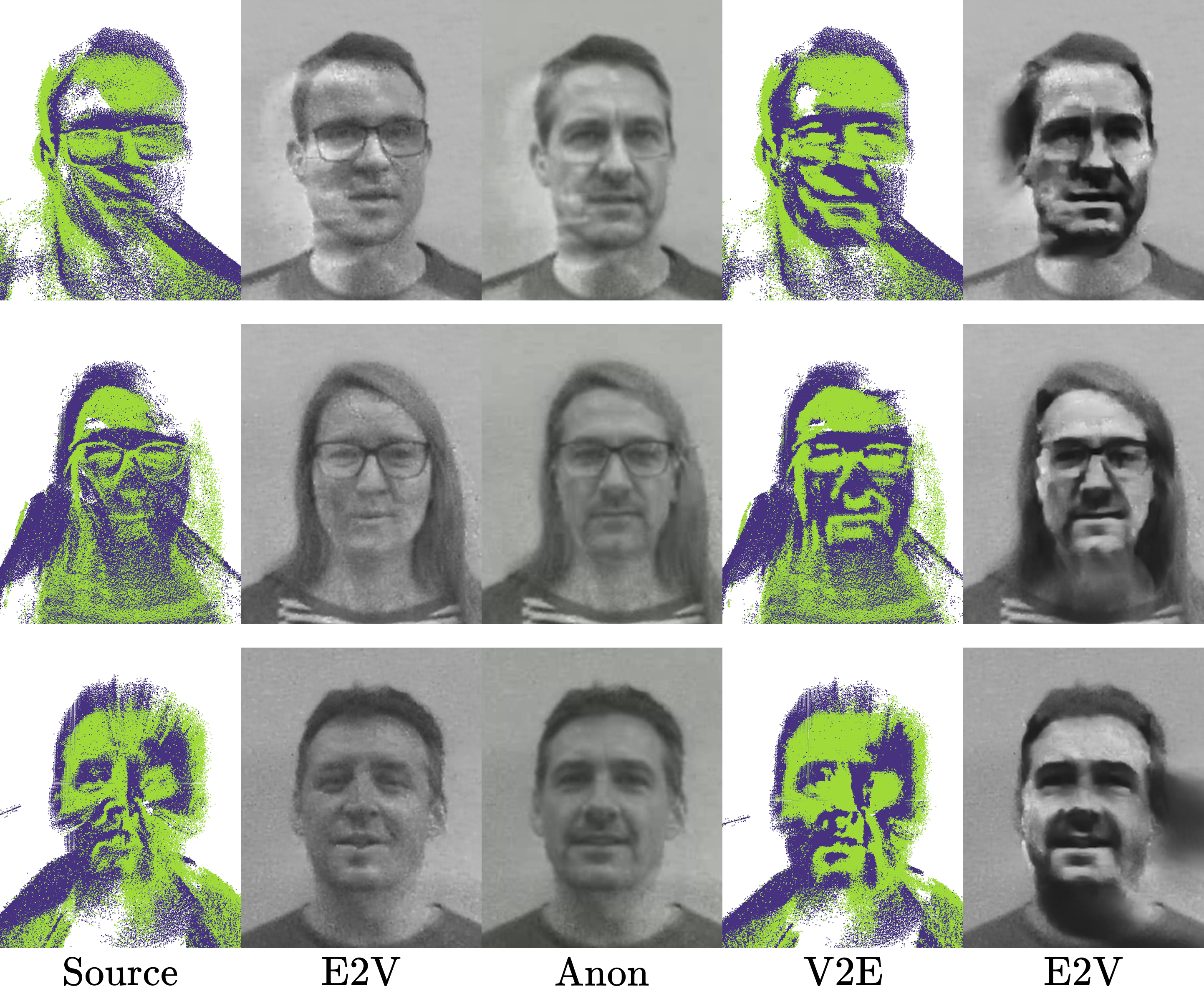}
   \caption{\textbf{Qualitative examples of source and synthetic identities.} Comparison of three subjects (rows). Columns from left-to-right: Source event streams, intermediate E2V representations, the anonymized generative output (Anon), V2E projection into a new event stream, and the final downstream E2V validation.}
   \label{fig:ExamplesQualit}
\end{figure}

\subsubsection{Qualitative}
Visual evaluation of the proposed pipeline demonstrates its capability to synthesize realistic, alternative identities in the event space (see~\cref{fig:ExamplesQualit}). To ensure a fair comparison, the conditioning command for the SD2 reference identity was kept constant across the shown examples. The generative anonymization successfully captures macroscopic facial mimicry, preserving essential expressions (e.g.,~\cref{fig:ExamplesQualit}, row three). However, the model occasionally struggles to translate subtle visual micro-expressions, particularly around the mouth region (e.g.,~\cref{fig:ExamplesQualit}, row two). Furthermore, we observe artifacts in the frame-space stemming from an imperfect spatiotemporal merge. Noticeable regions experience \textit{smearing}, tracking the movement of the synthetic faces. This is particularly evident when comparing the initial E2V step (\cref{fig:ExamplesQualit}, second column) with the final E2V validation step (\cref{fig:ExamplesQualit}, last column), where large zones of smeared black pixels are visible adjacent to the synthetic face.

\begin{table*}[t]
\centering
\small
\setlength{\tabcolsep}{4pt}
\caption{\textbf{Anonymization metrics and feature preservation in image space.} \textit{Anonymization} compares the source stream to our generative output. \textit{Reference} compares two independent source streams of the same subject, establishing the natural variance limit for identity and pose.}
\label{tab:AnonMetrics}
\resizebox{\textwidth}{!}{
\begin{tabular}{lcccccccc}
\toprule
\textbf{Metric} & \multicolumn{2}{c}{\textbf{Identity Similarity} $\downarrow$} & \multicolumn{2}{c}{\textbf{Temporal Stability} $\uparrow$} & \multicolumn{2}{c}{\textbf{Pose Error} $\downarrow$ / \textdegree} & \multicolumn{2}{c}{\textbf{Mimicry Error} $\downarrow$} \\
\cmidrule(lr){2-3} \cmidrule(lr){4-5} \cmidrule(lr){6-7} \cmidrule(lr){8-9}
 & Anonymization & Reference & Anonymization & Reference & Anonymization & Reference & Anonymization & Reference \\
\midrule
{\textbf{Mean $\mu$}} & 0.118 & 0.713 & 0.760 & 0.770 & 3.304 & 2.613 & 0.181 & 0.239 \\
{\textbf{Std. Deviation $\sigma$}} & 0.0172 & 0.0159 & 0.0135 & 0.0325 & 0.453 & 0.566 & 0.0298 & 0.1474 \\
\bottomrule
\end{tabular}
}
\end{table*}

\begin{table*}[t]
\centering
\small
\setlength{\tabcolsep}{4pt}
\caption{\textbf{Downstream task performance for face detection.} Intensity-domain (\textit{YOLO}) and event-domain (\textit{Event}) metrics are computed using YOLOv8~\cite{terven_comprehensive_2023} and an event-based detector~\cite{bissarinova_faces_2024}, respectively. IoU and detection rate error compare unmodified source data (\textit{Reference}) against the anonymized output (\textit{Anonymization}).}
\label{tab:DownstreamMetrics}
\resizebox{\textwidth}{!}{
\begin{tabular}{lccccc}
\toprule
\textbf{Metric} & \textbf{YOLO Conf. Anonymization} $\uparrow$ & \textbf{YOLO Conf. Reference} $\uparrow$ & \textbf{YOLO IoU} $\uparrow$ & \textbf{YOLO Det.-Rate Error} $\downarrow$ & \textbf{Event IoU} $\uparrow$ \\
\midrule
{\textbf{Mean $\mu$}} & 0.894 & 0.937 & 0.960 & 0.000 & 0.702 \\
{\textbf{Std. Deviation $\sigma$}} & 0.011 & 0.007 & 0.010 & 0.000 & 0.137 \\
\bottomrule
\end{tabular}
}
\end{table*}

\begin{table}[t]
\centering
\small
\setlength{\tabcolsep}{4pt}
\caption{\textbf{Structural anonymization metrics in the event space.} \textit{Anonymization} compares source and synthetic streams, while \textit{Reference} evaluates two independent captures of the same subject to provide a baseline geometric distance.}
\label{tab:EventMetrics}
\begin{tabular}{llcc}
\toprule
 & & \textbf{Mean $\mu$} & \textbf{Std. Deviation $\sigma$} \\
\midrule
{\textbf{STCD} $\uparrow$} & Anonymization & 0.3143 & 0.0415 \\
 & Reference & 0.0099 & 0.0029 \\
\cmidrule(lr){2-4}
{\textbf{EMD} $\uparrow$} & Anonymization & 0.1276 & 0.0132 \\
 & Reference & 0.0085 & 0.0039 \\
\bottomrule
\end{tabular}
\end{table}

\subsubsection{Quantitative}

\paragraph{Effectiveness of the Anonymization.} The primary goal of our framework is to obscure the original identity while maintaining the structural integrity of the event stream. As shown in~\cref{tab:AnonMetrics}, the identity similarity drops significantly from a baseline mean of 0.713 to 0.118 following anonymization.

This is corroborated by our proposed event-space metrics STCD and EMD (\cref{tab:EventMetrics}). When comparing two different identities, in the best case the fundamental facial structure and thus event generation differ. As a specific face generates a unique 3D structural topography in the event-space, differing identities force the nearest-neighbor search in STCD to reach further. As shown in~\cref{tab:EventMetrics}, we measure an increase in STCD from a baseline comparison to the anonymized data of $>$31 times, indicating a clear differentiation in facial structure. Alongside this increase, we also measure a clear increase in EMD from 0.0085 (reference) to 0.1276 (anonymized).

Together, these metrics indicate a successful global distributional shift and structural anonymization.

\paragraph{Spatio-Temporal Utility and Feature Preservation.} A core advantage of generative anonymization is the preservation of data utility. \Cref{tab:AnonMetrics} demonstrates that the measure for temporal stability remains virtually unchanged (baseline 0.760 to anonymized 0.770), indicating that the generated identity remains consistent across the full data stream. 

When evaluating pose and mimicry, it is important to note that the intra-subject reference compares two independent recordings, naturally capturing behavioral variance in head movement and facial expressions. Against this strict standard, the geometric alignment between the source video and its anonymized counterpart is highly preserved, yielding a pose error of just 3.304\textdegree. Notably, our method achieves a mimicry error of 0.181, which is tighter than the natural intra-subject variance of 0.239. This demonstrates, that the generative pipeline faithfully transfers the source facial expressions without introducing unintended synthetic deviations.

\paragraph{Downstream Task Performance.} We validate the utility of the anonymized event streams for practical applications using YOLOv8~\cite{redmon_you_2016,terven_comprehensive_2023} and an event-based detector~\cite{bissarinova_faces_2024} (\cref{tab:DownstreamMetrics}). The anonymization process introduces zero degradation to the overall detection capability, yielding no YOLO detection-rate discrepancy between anonymized and baseline data streams. The mean YOLO confidence score remains highly robust at 0.894, compared to the baseline of 0.937. Furthermore, the spatial bounding boxes remain tightly aligned, achieving a YOLO Intersection over Union (IoU) of 0.960 in grayscale space and an Event IoU of 0.702, confirming that the macroscopic spatiotemporal structure required for downstream perception tasks is fully preserved.

\vspace{0.13cm}
In summary, our quantitative and qualitative evaluations collectively demonstrate that the proposed generative framework successfully anonymizes subjects identities while fully preserving the essential spatio-temporal structure, facial mimicry, and downstream usability of the event stream.

\section{Limitations}
\label{sec:limitations}

\begin{figure}[t]
  \centering
   \includegraphics[width=\linewidth]{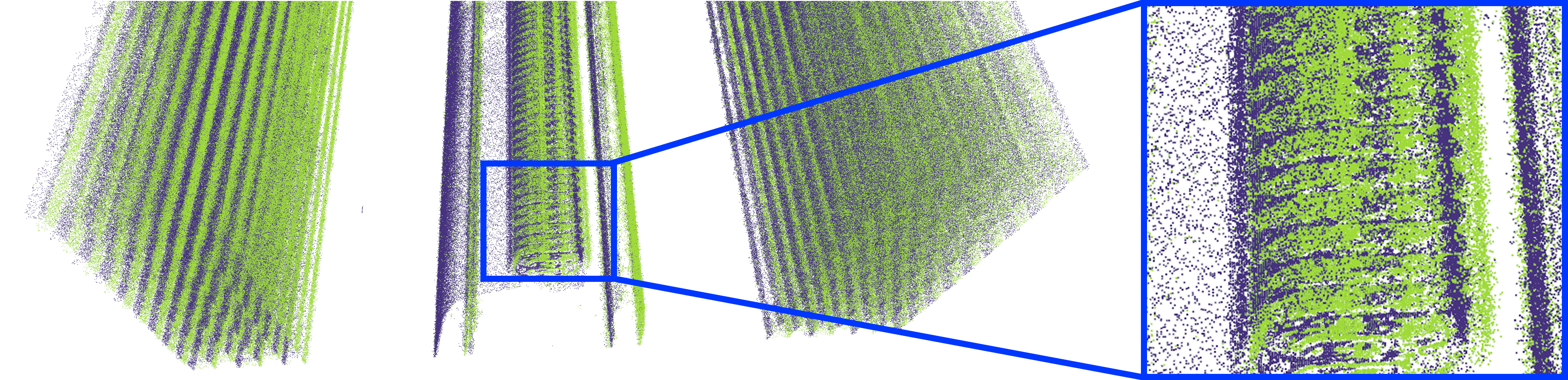}
   \caption{\textbf{V2E discretization and density artifacts.} Viewed tilted from the top-down position, where more recent events are closer to the frontal cross-section. The reverse projection step relies on standard V2E conversion, which leads to discretization in the event-space. Notably only in the parts of the event stream where information has been replaced (facial region).}
   \label{fig:example_v2e}
\end{figure}

While this work serves as a pioneering proof-of-concept for generative anonymization in the neuromorphic domain, it naturally presents several avenues for future refinement.

\begin{itemize}
    \item \textbf{Reliance on Frame-Based Intermediaries.} To bridge the current modality gap, our pipeline translates asynchronous event data into discrete grayscale frames to leverage established, high-fidelity models. Consequently, we do not yet directly alter the raw event stream natively. While recent destructive obfuscation methods operate directly on event spikes (e.g.,~\cite{ahmad_person_2023,ahmad_event_2024}), developing a native, spatiotemporal generative model that executes identity replacement directly on asynchronous neuromorphic data remains an open and highly challenging objective for future work.
    \item \textbf{Event Simulator Constraints and Density.} Our reverse projection step relies on standard video-to-event conversion, leading to discretization in the event-space (see~\cref{fig:example_v2e}). While continuous-time simulators like V2CE~\cite{zhang_v2ce_2024} theoretically offer a more native event representation, our experiments revealed that they currently fail to generate a sufficient density of events to adequately fill the high-frequency facial region. This sparse generation exacerbates smearing artifacts when the data is subsequently subjected to downstream E2V methods.
    \item \textbf{Resolution and Micro-Expression Bottlenecks.} The quality of the anonymized output is inherently upper-bounded by the specific face-swapping prior utilized in the pipeline. Currently, subtle visual micro-expressions can occasionally be lost during the translation process. Integrating more advanced, natively high-resolution diffusion models, such as DreamID~\cite{ye_dreamid_2025}, could enhance the fidelity of localized mimicry and dynamic facial details.
\end{itemize}

\section{Conclusion}
\label{sec:conclusion}

In this paper, we introduced the first generative anonymization framework for the event domain, successfully resolving the severe utility-privacy trade-off inherent in neuromorphic vision. By bridging the modality gap via intermediate intensity representations, our pipeline leverages high-fidelity models to seamlessly replace sensitive facial features with synthesized, alternative identities.

Evaluations demonstrate that our method reliably prevents identity recovery, while preserving the essential spatiotemporal structure, facial mimicry, and downstream task utility of the original event stream. To facilitate rigorous evaluation, we presented a novel, synchronized RGB-Event dataset, establishing a robust benchmark for future research. While native, asynchronous event-level generation remains an exciting open challenge, this work provides a critical foundation for the safe and privacy-preserving deployment of event cameras in public, human-centric environments.

{
    \small
    \bibliographystyle{ieeenat_fullname}
    \bibliography{main}
}

\end{document}